\documentclass[11pt,a4paper]{article}

\usepackage[utf8]{inputenc}
\usepackage[margin=1in]{geometry}
\usepackage{amsmath,amssymb,amsfonts,amsthm}
\usepackage{booktabs}
\usepackage{multirow}
\usepackage{graphicx}
\usepackage{adjustbox}
\usepackage{tikz}
\usetikzlibrary{shapes,arrows,positioning,calc,arrows.meta}
\usepackage{caption}
\usepackage{subcaption}
\usepackage[numbers,sort&compress]{natbib}
\usepackage{url}
\usepackage{microtype}
\usepackage{hyperref}
\hypersetup{
    colorlinks=true,
    linkcolor=blue!70!black,
    citecolor=blue!70!black,
    urlcolor=blue!70!black
}

\title{\textbf{Orthogonal Boundary Unlearning for Dual-Use Concepts in Large Language Models}}

\author{
  \textbf{Niccol\`o Ronchetti}
}
\date{}

\begin{document}

\maketitle

\begin{abstract}
Mitigating catastrophic risks from large language models requires excising hazardous operational capabilities, such as cyberwarfare exploitation and pathogen enhancement, without degrading benign scientific utility. In these dual-use domains, legitimate defensive science and malicious attacks share vocabulary, syntax, and internal model representations, causing standard unlearning methods to induce severe collateral damage (concept bleeding).

In this work, we present \textbf{Orthogonal Boundary Unlearning (OBU)}, a mathematically grounded framework designed to eliminate collateral damage on benign dual-use capabilities. OBU extracts a domain-matched capability subspace from paired calibration queries via singular value decomposition on empirical activation covariances, and restricts parameter updates strictly to the orthogonal null space. This guarantees that benign representations experience zero parameter perturbation along the spanned calibration subspace.

We present two complementary realizations of OBU: \textbf{Projected LoRA}, which applies the orthogonal constraint during iterative gradient-based fine-tuning, and \textbf{Closed-Form Weight Surgery}, which directly edits model weights in a single analytical step without training. Across four modern open-weight foundation models (Gemma-3-4B, Qwen3.5-4B, SmolLM3-3B, and Granite-4.2-3B), OBU achieves strong dual-use concept separation, substantially mitigates in-context priming attacks, and demonstrates durable resistance against adversarial relearning without degrading benign domain competence.
\end{abstract}

\section{Introduction}
\label{sec:intro}

Large language models (LLMs) trained on web-scale text corpora acquire extensive domain expertise spanning both transformative scientific disciplines and hazardous technologies~\citep{li2024wmdp,qwen2026,gemma3}. A central challenge in artificial intelligence safety is mitigating risks from \emph{dual-use capabilities}: domains where legitimate scientific utility and catastrophic hazards are deeply intertwined~\citep{conceptguard2026,peleg2026forget}. Dual-use dilemmas emerge prominently in: (i) cybersecurity, where defensive firmware analysis, vulnerability patch verification, and intrusion detection share foundations with automated zero-day exploitation; (ii) computational biology and virology, where pathogen vaccine candidate modeling and therapeutic antibody design share generative mechanics with viral transmissibility enhancement; (iii) molecular chemistry, where industrial synthesis optimization mirrors the generation of schedule-one toxic precursors; and (iv) nuclear engineering, where civilian power reactor modeling, medical radioisotope production, and neutron transport simulations share theoretical foundations with proliferation-sensitive fissile enrichment and weaponization physics~\citep{li2024wmdp,liu2024rethinking}.

\begin{figure*}[t]
\centering
\begin{tikzpicture}[scale=1.05, >=Stealth]
  % Legend / Title positioned in the upper-left margin
  \node[font=\normalsize\bfseries, text=black!75, anchor=south west] at (-6.2, 2.9) {Manifold $\mathcal{M} \subset \mathbb{R}^d$};

  % Category header labels positioned above the respective regions
  \node[font=\small\bfseries, text=red!85!black, anchor=south] at (-3.4, 2.2) {$\mathcal{M}_{\text{hazard}}$};
  \node[font=\small\bfseries, text=black!75, anchor=south] at (0.0, 2.2) {Boundary $\mathcal{B}$};
  \node[font=\small\bfseries, text=blue!85!black, anchor=south] at (3.5, 2.2) {$\mathcal{M}_{\text{benign}}$};

  % 1. Wide, Deeply Curved 3D Manifold Surface M (Dramatic Saddle Curvature)
  \shade[left color=red!18, right color=blue!18, draw=black!75, line width=1.1pt]
    (-6.2, -0.6) .. controls (-2.5, -2.8) and (2.2, -0.4) .. (6.2, -0.8)
    -- (5.8, 1.8) .. controls (2.2, 2.7) and (-2.5, 0.4) .. (-5.8, 1.6)
    -- cycle;

  % Longitudinal 3D contour curves emphasizing the dramatic saddle depth
  \draw[black!25, thin] 
    (-6.0, 0.5) .. controls (-2.5, -1.2) and (2.2, 1.2) .. (6.0, 0.5);
  \draw[black!25, thin] 
    (-6.1, -0.1) .. controls (-2.5, -2.0) and (2.2, 0.4) .. (6.1, -0.2);

  % Transverse contour curves down the dip
  \draw[black!20, thin] 
    (-3.8, 1.3) .. controls (-3.9, 0.1) .. (-3.8, -1.2);
  \draw[black!20, thin] 
    (3.8, 1.6) .. controls (3.9, 0.4) .. (3.8, -0.8);

  % 2. Operational Safety Boundary B (Dashed line down the central dip)
  \draw[dashed, line width=1.3pt, black!65] 
    (0.0, 1.7) .. controls (-0.25, 0.3) and (0.15, -0.9) .. (0.0, -1.8);

  % 3. Hazardous Activations on M_hazard (Left Wing)
  \fill[red!85] (-4.5, 0.9) circle (2.8pt) node[below=2pt, font=\scriptsize\bfseries, text=red!90!black] {$h_f^{(1)}$};
  \fill[red!85] (-2.8, 0.3) circle (2.8pt) node[below=2pt, font=\scriptsize\bfseries, text=red!90!black] {$h_f^{(2)}$};
  \fill[red!85] (-3.8, -0.6) circle (2.8pt) node[below=2pt, font=\scriptsize\bfseries, text=red!90!black] {$h_f^{(3)}$};

  % 4. Benign Activations on M_benign (Right Wing)
  \coordinate (B1) at (2.2, 0.3);
  \coordinate (B2) at (3.6, 0.8);
  \coordinate (B3) at (4.8, 0.0);
  \fill[blue!85] (B1) circle (2.8pt) node[below=2pt, font=\scriptsize\bfseries, text=blue!90!black] {$h_b^{(1)}$};
  \fill[blue!85] (B2) circle (2.8pt) node[below=2pt, font=\scriptsize\bfseries, text=blue!90!black] {$h_b^{(2)}$};
  \fill[blue!85] (B3) circle (2.8pt) node[below=2pt, font=\scriptsize\bfseries, text=blue!90!black] {$h_b^{(3)}$};

  % 5. Linear Subspace V_boundary as a Clean Rectangle
  \draw[fill=blue!12, draw=blue!80, line width=1.2pt, fill opacity=0.7]
    (1.4, -0.7) rectangle (5.6, 1.4);
  \node[font=\small\bfseries, text=blue!95!black, anchor=north] at (3.5, -1.35) {$\mathcal{V}_{\text{boundary}} = \operatorname{span}(V_r)$};
\end{tikzpicture}
\caption{\textbf{Dual-Use Representation Manifold and Benign Subspace Identification.} In language model activation space $\mathbb{R}^d$, capabilities reside on a continuous representation manifold $\mathcal{M}$ partitioned by an operational safety boundary $\mathcal{B}$ into hazardous capabilities $\mathcal{M}_{\text{hazard}}$ (activations $\{h_f^{(i)}\}$, where subscript $f$ denotes the forget set) and benign science $\mathcal{M}_{\text{benign}}$ (activations $\{h_b^{(i)}\}$). Because dual-use pairs lie in close proximity across $\mathcal{B}$, isolating benign capabilities requires identifying their local geometry. OBU passes domain-matched calibration queries $\mathcal{D}_b$ through the model to extract benign activations $\{h_b^{(i)}\}$, computing the principal linear subspace $\mathcal{V}_{\text{boundary}} = \operatorname{span}(V_r)$ spanned by their leading $r$ empirical singular vectors to tightly enclose the local benign capability manifold.}
\label{fig:activation_space_clustering}
\end{figure*}
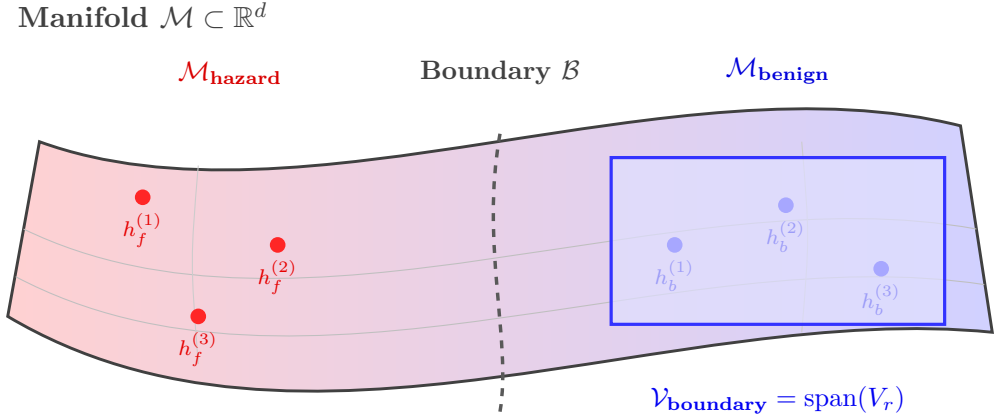

When safety practitioners attempt to excise dangerous capabilities via standard machine unlearning techniques—such as gradient ascent (GA), Representation Misdirection for Unlearning (RMU)~\citep{li2024wmdp}, or Negative Preference Optimization (NPO)~\citep{simnpo2024}—they confront the pathology of \emph{concept bleeding}~\citep{zhang2024easing,conceptguard2026,hu2025blur}. Because hazardous and benign dual-use tasks share vocabulary, syntactic constructs, and reasoning pathways, unconstrained gradient updates propagate across shared weight matrices and degrade benign scientific competence~\citep{liu2024rethinking,preunlearn2026}.

\subsection{The Geometry of Dual-Use Entanglement}
Dual-use capability entanglement is an intrinsic geometric property of transformer representations driven by three structural factors:
\begin{enumerate}
    \item \textbf{Extensive Lexical and Syntactic Overlap:} In dual-use domains, benign scientific workflows and hazardous operations share identical technical vocabulary, syntactic constructs, and functional terminology (such as low-level system calls, chemical reaction equations, or genetic sequence notations). Because autoregressive pre-training optimizes next-token prediction across these shared lexical distributions, the model embeds benign and hazardous prompts into shared initial representation spaces.
    \item \textbf{Early-to-Mid Layer Representation Alignment:} In transformer architectures, early and intermediate feed-forward and attention layers ($l \in \{1, \dots, L/2\}$) construct general conceptual representations, token associations, and syntactical bindings~\citep{geva2021transformer,park2023linear}. In these early-to-mid representations, dual-use pairs share substantial subspace structure, with task-specific operational trajectories diverging predominantly in deeper transformer layers.
    \item \textbf{Parametric Superposition:} Deep networks store far more features than available dimensions through \emph{polysemantic superposition}~\citep{elhage2022toy,henighan2023superposition}. Neurons and attention heads are shared across orthogonal concepts, so perturbing weight matrices without subspace constraints collides with adjacent factual associations.
\end{enumerate}

We emphasize that our designation of paired contrastive tasks as residing at the ``boundary'' is an operational, qualitative characterization based on prompt construction, rather than an explicitly measured metric margin in activation space $\mathbb{R}^{d_{\text{in}}}$. As illustrated in Figure~\ref{fig:activation_space_clustering}, because paired dual-use queries differ primarily in terminal generative intent (e.g., defensive patch verification versus weaponized exploitation), their activations $\{h_f^{(i)}\}$ and $\{h_b^{(i)}\}$ lie in close proximity across the operational safety boundary $\mathcal{B}$ on the representation manifold $\mathcal{M}$.

To circumvent collateral damage, recent works propose null-space unlearning or representation erasure, such as Null-Space Constrained Low-Rank Adaptation (NSRU)~\citep{nsru2026} and ZeroUnlearn~\citep{zerounlearn2026}. However, these approaches compute retain spaces globally over generic web text or broad instruction datasets. While such global null spaces protect generic capabilities (such as trivia or common coding), they fail to isolate the local benign region $\mathcal{M}_{\text{benign}}$ adjacent to $\mathcal{B}$, leaving benign dual-use capabilities vulnerable to unlearning updates.

\subsection{Contributions}
To resolve dual-use capability entanglement without collateral damage, we introduce \textbf{Orthogonal Boundary Unlearning (OBU)}, an unlearning framework that extracts the local benign subspace $\mathcal{V}_{\text{boundary}} = \operatorname{span}(V_r)$ (Figure~\ref{fig:activation_space_clustering}) and restricts parameter updates strictly to its orthogonal complement (Figure~\ref{fig:subspace_projection_mechanisms}):

\begin{itemize}
    \item \textbf{Boundary Subspace Projection:} We construct layer-wise orthogonal projectors $P_{\text{boundary}}^{(l)\perp} = I - V_{l,r} V_{l,r}^T$ from the uncentered token-level second-moment matrix $K_{\text{boundary}}^{(l)}$ of paired benign calibration queries, guaranteeing zero perturbation along $\mathcal{V}_{\text{boundary}}^{(l)} = \operatorname{span}(V_{l,r})$ and bounding residual variance by $1-\tau$.
    \item \textbf{Subspace Capacity and Layer Profiling:} Across four open-weight models, we show that isolating the benign capability subspace $\mathcal{V}_{\text{boundary}}$ does not exhaust model capacity: in the targeted final $25\%$ of attention output layers ($o\_proj$), capturing $95\%$ of benign activation variance requires on average only $8.0\%$ of the layer dimensions, leaving $92.0\%$ of the dimensions available in the orthogonal complement for unlearning (nearly twice as compact as empirical Fisher information at $16.2\%$, while avoiding the $16\times\text{--}25\times$ covariance memory overhead of high-dimensional MLP projections). Even under a strict $99\%$ variance threshold across large calibration sets, $27\%\text{--}73\%$ of the ambient dimensions remain available in the orthogonal null space.
    \item \textbf{Complementary Algorithmic Realizations:} We instantiate OBU in two algorithms that fold directly into base weights with zero inference overhead (reported below at $\tau=0.95$):
    \begin{itemize}
        \item \emph{Projected LoRA} filters adapter inputs through $P_{\text{boundary}}^{(l)\perp}$ during gradient ascent on hazardous text, suppressing hazardous compliance on paired dual-use queries from $22.5\%$ down to $3.9\%$ on Qwen3.5-4B (and from $30.6\%$ to $23.3\%$ on Gemma-3-4B-IT) while preserving $100\%$ accuracy on matched benign queries.
        \item \emph{Closed-Form Weight Surgery} analytically subtracts the principal hazardous subspace within the benign null space ($W_l - W_l P_f^{(l)} P_{\text{boundary}}^{(l)\perp}$) in a single training-free step, reducing multiple-choice hazardous accuracy on WMDP-Cyber from $41.1\%$ to $39.1\%$ on Gemma and from $38.4\%$ to $31.5\%$ on Granite while maintaining $\ge 99.6\%$ accuracy on matched benign queries.
    \end{itemize}
    \item \textbf{Robustness to In-Context Priming and Adversarial Relearning:} We evaluate whether unlearned capabilities can be re-elicited via in-context priming (prepending legitimate scientific demonstrations to hazardous requests) or adversarial fine-tuning on held-out hazardous text. On Qwen3.5-4B, Projected LoRA cuts direct 0-shot harmful compliance from $17.9\%$ to $1.5\%$ and 1-shot primed compliance from $37.4\%$ to $15.8\%$. Furthermore, whereas standard unconstrained LoRA re-acquires $50\%$ of its unlearned capability within $30$ fine-tuning steps, both OBU realizations resist recovery across the entire 100-step fine-tuning horizon.
\end{itemize}

\section{Related Work}
\label{sec:related_work}

\subsection{Machine Unlearning and Dual-Use Collateral Damage}
Machine unlearning originated in classification to remove the influence of specific training samples without full retraining~\citep{bourtoule2021machine,nguyen2022survey}, and has since been adapted to LLMs to erase memorized text or hazardous capabilities~\citep{eldan2023whoshp,jang2022knowledge,yao2024unlearning}. Standard optimization methods—including Gradient Ascent (GA), Representation Misdirection for Unlearning (RMU)~\citep{li2024wmdp}, and preference-based objectives such as DPO~\citep{rafailov2023direct} and Negative Preference Optimization (NPO)~\citep{simnpo2024}—penalize hazardous outputs directly across model weights. While effective when the target knowledge is isolated from general capabilities (such as forgetting fictitious author biographies in TOFU~\citep{tofu2024}), unconstrained updates in dual-use domains alter shared representations and induce severe \emph{concept bleeding} (over-forgetting) on adjacent benign tasks~\citep{liu2024rethinking,peleg2026forget}. Recent benchmarks specifically designed to measure forget-retain overlap—including ConceptGuard~\citep{conceptguard2026}, BLUR~\citep{hu2025blur}, and PreUnlearn~\citep{preunlearn2026}—confirm that standard unlearning objectives struggle to suppress hazardous responses without degrading closely related benign expertise.

\subsection{Subspace Projection and Null-Space Adaptation}
Projecting updates onto orthogonal subspaces was originally developed in continual learning to prevent catastrophic forgetting across sequential tasks~\citep{zeng2019continual,farajtabar2020orthogonal} and in linear concept erasure to remove protected attributes from representations (INLP~\citep{ravfogel2020null}, LEACE~\citep{leace2023,ravfogel2022linear}). In parameter-efficient fine-tuning, O-LoRA~\citep{wang2023olora} penalizes overlap between sequential task adapters ($A_i^T A_t \approx 0$), while Safe LoRA~\citep{hsu2024safelora} projects fine-tuned adapters onto the weight difference ($W_{\text{aligned}} - W_{\text{base}}$) to preserve safety guardrails during downstream fine-tuning. However, neither method can unlearn pre-trained capabilities or isolate entangled concepts within a single base model.

Most closely related to our approach, recent unlearning frameworks such as NSRU~\citep{nsru2026}, ZeroUnlearn~\citep{zerounlearn2026}, and CIR~\citep{sondej2025cir} adapt activation null-space projection to LLM unlearning. While these works constrain updates using generic retain corpora or supervised refusal targets, they do not address the severe representation overlap of paired dual-use tasks—where a critical open question is whether projecting out domain-matched benign activations leaves sufficient null-space capacity for unlearning. OBU directly addresses this regime by constructing $P_{\text{boundary}}^\perp$ from paired contrastive boundary queries, profiling spectral capacity across module types and depths, and comparing iterative (Projected LoRA) and training-free (Closed-Form Surgery) realizations under in-context priming and relearning attacks.

\subsection{Adversarial Stress-Testing and Mechanistic Auditing}
Recent audits demonstrate that suppressing accuracy on multiple-choice benchmarks such as WMDP~\citep{li2024wmdp} does not guarantee genuine capability removal~\citep{lynch2024eight,wmdp_plus_plus2026}. Unlearned models frequently re-exhibit hazardous capabilities when subjected to optimization-based jailbreaks~\citep{zou2023universal,pair2023,liu2023autodan,mehrotra2023tree,schwinn2024softprompt}, in-context priming with benign or cooperative demonstrations~\citep{wei2023jailbroken,anil2024many,jailbreakbench2024,bhatt2023purple}, or lightweight fine-tuning on a small number of hazardous examples (relearning attacks)~\citep{henderson2023selfdestruct,tamirisa2024unlearning,jaccess2026}. To evaluate whether unlearning alters internal circuits or merely induces shallow refusals, mechanistic interpretability studies employ Sparse Autoencoders (SAEs)~\citep{bricken2023monosemanticity,cunningham2023sparse,lieberum2024gemma}, linear representation probes~\citep{marks2023geometry,park2023linear,belrose2023tuned}, and gradient sensitivity analysis. We combine paired dual-use evaluation on ConceptGuard~\citep{conceptguard2026} with multi-shot priming attacks and 100-step fine-tuning relearning audits to evaluate OBU's robustness.

\section{Mathematical Formulation of OBU}
\label{sec:method}

This section formalizes the Orthogonal Boundary Unlearning (OBU) framework. We define the empirical boundary covariance, derive the truncated singular value decomposition (SVD) rank selection, establish the asymmetric defense mechanism, and present two complementary algorithmic realizations: Projected LoRA and Closed-Form Weight Surgery.

\subsection{Empirical Boundary Covariance and Subspace Decomposition}
Consider a transformer language model parameterized by weights $W$, with a targeted linear projection layer $W_l \in \mathbb{R}^{d_{\text{out}} \times d_{\text{in}}}$ at layer $l \in \{1, \dots, L\}$. During autoregressive processing of a sequence of length $T$, the module receives hidden activation vectors $h_{l,t} \in \mathbb{R}^{d_{\text{in}}}$ at each token position $t \in \{1, \dots, T\}$ and produces intermediate output states $z_{l,t} = W_l h_{l,t} \in \mathbb{R}^{d_{\text{out}}}$.

To construct the projection operator, we curate a calibration dataset $\mathcal{D}_b = \{x_b^{(i)}\}_{i=1}^M$ of $M$ domain-matched benign queries selected as contrastive counterparts to the hazardous domain. Passing each query $x_b^{(i)}$ of token length $T_i$ through the model, we extract the input activations $\{h_{l,t}(x_b^{(i)})\}_{t=1}^{T_i}$ immediately preceding $W_l$ across all token positions.

Following prior null-space projection methods~\citep{zeng2019continual,nsru2026,zerounlearn2026}, we accumulate the uncentered empirical second-moment matrix (referred to hereafter as the empirical boundary covariance matrix) $K_{\text{boundary}}^{(l)} \in \mathbb{R}^{d_{\text{in}} \times d_{\text{in}}}$ across all $N_{\text{tok}} = \sum_{i=1}^M T_i$ calibration tokens:
\begin{equation}
K_{\text{boundary}}^{(l)} = \frac{1}{N_{\text{tok}}}\sum_{i=1}^M \sum_{t=1}^{T_i} h_{l,t}(x_b^{(i)}) \left(h_{l,t}(x_b^{(i)})\right)^T
\label{eq:k_boundary}
\end{equation}
Using the uncentered second moment rather than a mean-centered covariance is mathematically required because linear layers $z_{l,t} = W_l h_{l,t}$ act directly on uncentered activations $h_{l,t}$; ensuring $\Delta W_l h_{l,t} \approx \mathbf{0}$ requires projecting out the uncentered span of $h_{l,t}$ (including its mean direction). Accumulating across all $N_{\text{tok}}$ tokens rather than a single token per prompt enables rank estimation up to $\min(N_{\text{tok}}, d_{\text{in}})$ even when $M < d_{\text{in}}$; although this includes common function words alongside domain terms, recurring syntactic directions are absorbed into the leading singular vectors (protecting general language fluency), while in deep causal layers each token state $h_{l,t}$ already integrates the preceding query context.

Because $K_{\text{boundary}}^{(l)}$ is symmetric and positive semi-definite, its singular value decomposition coincides with its eigendecomposition:
\begin{equation}
K_{\text{boundary}}^{(l)} = U_l \Sigma_l V_l^T = V_l \Sigma_l V_l^T
\label{eq:truncated_svd}
\end{equation}
where $V_l = [v_{l,1}, v_{l,2}, \dots, v_{l,d_{\text{in}}}] \in \mathbb{R}^{d_{\text{in}} \times d_{\text{in}}}$ is an orthonormal basis of singular vectors, and $\Sigma_l = \operatorname{diag}(\sigma_{l,1}, \dots, \sigma_{l,d_{\text{in}}})$ contains singular values sorted in descending order: $\sigma_{l,1} \ge \sigma_{l,2} \ge \dots \ge \sigma_{l,d_{\text{in}}} \ge 0$.

\subsection{Energy-Preserving Rank Selection}
The singular values $\sigma_{l,j}$ quantify the activation variance concentrated along principal direction $v_{l,j}$ across benign boundary queries. Rather than fixing an arbitrary rank $r$, we determine $r$ dynamically via a cumulative energy threshold $\tau \in (0, 1)$:
\begin{equation}
r(\tau) = \min \left\{ k \in \{1, \dots, d_{\text{in}}\} \;\middle|\; \frac{\sum_{j=1}^k \sigma_{l,j}}{\sum_{j=1}^{d_{\text{in}}} \sigma_{l,j}} \ge \tau \right\}
\label{eq:rank_selection}
\end{equation}
Let $V_{l,r} = [v_{l,1}, \dots, v_{l,r}] \in \mathbb{R}^{d_{\text{in}} \times r}$ denote the matrix of the top-$r$ singular vectors. The principal benign capability subspace at layer $l$ is defined as $\mathcal{V}_{\text{boundary}}^{(l)} = \operatorname{span}(V_{l,r})$. The orthogonal boundary projection operator $P_{\text{boundary}}^{(l)\perp} \in \mathbb{R}^{d_{\text{in}} \times d_{\text{in}}}$ is constructed as:
\begin{equation}
P_{\text{boundary}}^{(l)\perp} = I_{d_{\text{in}}} - V_{l,r} V_{l,r}^T
\label{eq:p_boundary_perp}
\end{equation}
By construction, $P_{\text{boundary}}^{(l)\perp}$ is symmetric and idempotent ($(P_{\text{boundary}}^{(l)\perp})^2 = P_{\text{boundary}}^{(l)\perp}$), representing an orthogonal projection onto the null space of the benign calibration subspace: $\mathcal{V}_{\text{boundary}}^{(l)\perp} = \ker(V_{l,r}^T)$.

\subsection{Orthogonal Non-Interference Guarantee}
\begin{figure*}[!t]
\centering
\begin{subfigure}[t]{0.95\textwidth}
  \centering
  \begin{tikzpicture}[scale=0.88, >=Stealth]
    % Wide Benign Subspace Plane (Full Horizontal Width)
    \draw[fill=blue!8, draw=blue!75, line width=1.1pt] 
      (-6.0, -1.4) -- (1.5, -2.1) -- (6.0, 0.4) -- (-1.5, 1.1) -- cycle;
    % Subspace label positioned high inside the plane, completely clear of the perimeter
    \node[font=\small\bfseries, text=blue!90!black, anchor=south west] at (-5.2, 0.2) {$\mathcal{V}_{\text{boundary}}$};

    % Origin
    \coordinate (O) at (0, -0.5);

    % Benign vector inside the subspace
    \coordinate (Hb) at (4.0, -0.2);
    \draw[->, line width=2pt, blue!90!black] (O) -- (Hb) 
      node[midway, above=2pt, font=\small\bfseries, text=blue!90!black] {$h_{\text{benign}}$};

    % Oblique Unconstrained Update Vector (Dominant Right Singular Direction)
    \coordinate (Update) at (2.2, 2.0);
    \draw[->, line width=2.4pt, purple!85] (O) -- (Update) 
      node[above=2pt, font=\small\bfseries, text=purple!95!black] {Unconstrained $\Delta W$};

    % Projection dropping onto the benign subspace
    \coordinate (Shadow) at (2.2, -0.5);
    \draw[dashed, line width=1.2pt, red!70] (Update) -- (Shadow);
    \draw[->, line width=2pt, red!85] (O) -- (Shadow) 
      node[midway, below=3pt, font=\small\bfseries, text=red!90!black] {$\operatorname{Proj}_{\mathcal{V}_{\text{boundary}}}(\Delta W) \neq \mathbf{0}$};
  \end{tikzpicture}
  \caption{\textbf{Standard Unlearning (Oblique Operator Alignment):} Because parameter updates $\Delta W$ are unconstrained in parameter space, the operator's principal input singular vector is tilted obliquely relative to the domain-matched benign subspace $\mathcal{V}_{\text{boundary}}$. The resulting projection onto the benign subspace is non-zero ($\operatorname{Proj}_{\mathcal{V}_{\text{boundary}}}(\Delta W) \neq \mathbf{0}$), causing collateral damage to benign capabilities.}
  \label{fig:subspace_standard}
\end{subfigure}

\vspace{0.25cm}

\begin{subfigure}[t]{0.95\textwidth}
  \centering
  \begin{tikzpicture}[scale=0.88, >=Stealth]
    % Identical Wide Benign Subspace Plane
    \draw[fill=blue!8, draw=blue!75, line width=1.1pt] 
      (-6.0, -1.4) -- (1.5, -2.1) -- (6.0, 0.4) -- (-1.5, 1.1) -- cycle;
    % Subspace label positioned high inside the plane, completely clear of the perimeter
    \node[font=\small\bfseries, text=blue!90!black, anchor=south west] at (-5.2, 0.2) {$\mathcal{V}_{\text{boundary}}$};

    % Origin
    \coordinate (O) at (0, -0.5);

    % Benign vector inside the subspace
    \coordinate (Hb) at (4.0, -0.2);
    \draw[->, line width=2pt, blue!90!black] (O) -- (Hb) 
      node[midway, above=2pt, font=\small\bfseries, text=blue!90!black] {$h_{\text{benign}}$};

    % Strictly 90° Vertical OBU Update Vector
    \coordinate (UpdateOrth) at (0, 2.2);
    \draw[->, line width=2.4pt, teal!90!black] (O) -- (UpdateOrth) 
      node[above=2pt, font=\small\bfseries, text=teal!90!black] {$\Delta W P_{\text{boundary}}^\perp$};

    % 3D Right-angle marker at origin
    \draw[thick, teal!85] (0, -0.2) -- (0.3, -0.15) -- (0.3, -0.45);

    % Projection onto plane collapses to zero at origin
    \fill[teal!90!black] (O) circle (3.5pt);
    \node[below=4pt, font=\small\bfseries, text=teal!95!black] at (O) 
      {$\operatorname{Proj}_{\mathcal{V}_{\text{boundary}}}(\Delta W P_{\text{boundary}}^\perp) = \mathbf{0}$};
  \end{tikzpicture}
  \caption{\textbf{OBU Orthogonal Insulation (Zero Projection):} Right-multiplication of $\Delta W$ by the orthogonal projector $P_{\text{boundary}}^\perp$ (projecting incoming activations before $\Delta W$ is applied) restricts the operator's co-image strictly to $\mathcal{V}_{\text{boundary}}^\perp$, ensuring that $\mathcal{V}_{\text{boundary}} \subseteq \ker(\Delta W P_{\text{boundary}}^\perp)$. The projection onto the benign subspace vanishes ($\operatorname{Proj}_{\mathcal{V}_{\text{boundary}}}(\Delta W P_{\text{boundary}}^\perp) = \mathbf{0}$), mathematically guaranteeing zero perturbation of benign activations.}
  \label{fig:subspace_bcoa}
\end{subfigure}
\caption{\textbf{Operator Subspace Geometry and Orthogonal Projection in OBU.} Parameter updates $\Delta W \in \mathbb{R}^{d_{\text{out}} \times d_{\text{in}}}$ are depicted via their principal input singular direction in $\mathbb{R}^{d_{\text{in}}}$, where $\operatorname{Proj}_{\mathcal{V}_{\text{boundary}}}(\cdot)$ denotes the component of the update acting on the benign subspace $\mathcal{V}_{\text{boundary}}$ (its shadow on the blue plane).}
\label{fig:subspace_projection_mechanisms}
\end{figure*}
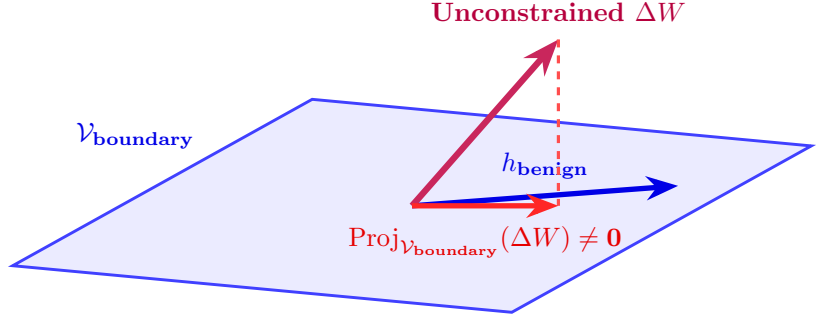
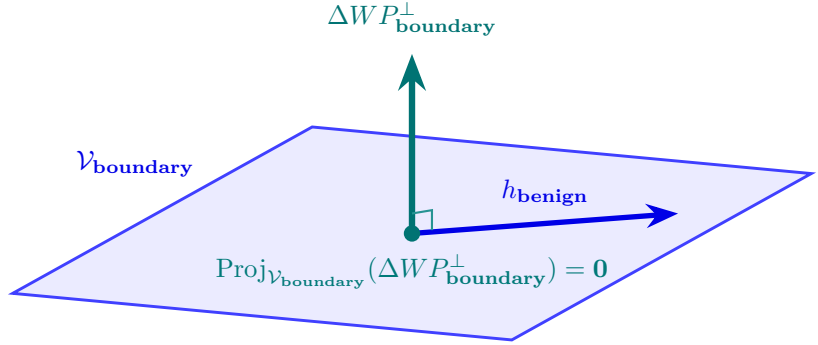

OBU separates hazardous and benign capabilities asymmetrically: benign boundary queries $\mathcal{D}_b$ define a hard geometric constraint ($P_{\text{boundary}}^{(l)\perp}$), while hazardous forget queries $\mathcal{D}_f$ (where subscript $f$ denotes the forget set) drive the weight update ($\Delta W_l$) within the remaining orthogonal complement. Consider any activation vector $h_b \in \mathcal{V}_{\text{boundary}}^{(l)}$. Expressing $h_b = V_{l,r} c$ for some coefficient vector $c \in \mathbb{R}^r$, we have:
\begin{equation}
P_{\text{boundary}}^{(l)\perp} h_b = \left(I_{d_{\text{in}}} - V_{l,r} V_{l,r}^T\right) V_{l,r} c = \left(V_{l,r} - V_{l,r} (V_{l,r}^T V_{l,r})\right) c = \mathbf{0}
\label{eq:asymmetric_defense}
\end{equation}
since $V_{l,r}^T V_{l,r} = I_r$. Consequently, for any weight update $\Delta W_l \in \mathbb{R}^{d_{\text{out}} \times d_{\text{in}}}$ right-multiplied by $P_{\text{boundary}}^{(l)\perp}$ (so that incoming activations are projected before $\Delta W_l$ is applied; Figure~\ref{fig:subspace_projection_mechanisms}), the perturbation along the benign boundary subspace is identically zero:
\begin{equation}
\left(\Delta W_l P_{\text{boundary}}^{(l)\perp}\right) h_b = \mathbf{0}, \quad \forall h_b \in \mathcal{V}_{\text{boundary}}^{(l)}
\end{equation}
More generally, any benign activation can be decomposed into its in-subspace projection and orthogonal residual as $h_b = h_{b,\parallel} + h_{b,\perp}$, where $h_{b,\parallel} \in \mathcal{V}_{\text{boundary}}^{(l)}$ and $h_{b,\perp} \in \mathcal{V}_{\text{boundary}}^{(l)\perp}$. The projected update then evaluates to $(\Delta W_l P_{\text{boundary}}^{(l)\perp}) h_b = \Delta W_l h_{b,\perp}$. Because the eigenvalues $\sigma_{l,j}$ of $K_{\text{boundary}}^{(l)}$ equal the mean squared projections $\mathbb{E}[\langle h_b, v_{l,j}\rangle^2]$ over the calibration tokens, a cumulative energy threshold $\tau$ guarantees that the expected residual energy satisfies $\mathbb{E}[\|h_{b,\perp}\|^2] = \sum_{j > r} \sigma_{l,j} \le (1-\tau)\,\mathbb{E}[\|h_b\|^2]$, bounding perturbations along the $(1-\tau)$ truncation tail.

\subsection{Realization 1: Projected LoRA}
In gradient-based adaptation, low-rank adaptation (LoRA)~\citep{lora2021} parameterizes weight updates as $\Delta W_l = \frac{\alpha}{k} B_l A_l$, where $B_l \in \mathbb{R}^{d_{\text{out}} \times k}$, $A_l \in \mathbb{R}^{k \times d_{\text{in}}}$, and $k \ll \min(d_{\text{in}}, d_{\text{out}})$ denotes the adapter intrinsic rank (distinguished from the SVD boundary truncation rank $r$). OBU Projected LoRA enforces boundary orthogonality by projecting the incoming activation $h$ before the low-rank adapter:
\begin{equation}
h_{\text{out}} = W_l h + \frac{\alpha}{k} B_l A_l \left(P_{\text{boundary}}^{(l)\perp} h\right)
\label{eq:projected_lora}
\end{equation}
During the backward pass, gradients with respect to adapter matrix $A_l$ are strictly projected onto $\mathcal{V}_{\text{boundary}}^{(l)\perp}$:
\begin{equation}
\nabla_{A_l} \mathcal{L} = \left(\frac{\alpha}{k} B_l^T \frac{\partial \mathcal{L}}{\partial h_{\text{out}}}\right) \left(P_{\text{boundary}}^{(l)\perp} h\right)^T
\end{equation}
so any gradient component residing in $\mathcal{V}_{\text{boundary}}^{(l)}$ is identically zeroed. After optimization, the adapter weights fold into the base matrix as $W_l' = W_l + \frac{\alpha}{k} B_l A_l P_{\text{boundary}}^{(l)\perp}$, incurring zero inference latency or parameter overhead.

We optimize the adapter parameters $\Theta = \{A_l, B_l\}$ over hazardous forget sequences $x \sim \mathcal{D}_f$ by maximizing the language modeling cross-entropy loss $\mathcal{L}_{\text{CE}}(x; \Theta) = -\frac{1}{T}\sum_{t=1}^T \log P_\Theta(x_t \mid x_{<t})$ via a smooth, bounded gradient ascent objective:
\begin{equation}
\mathcal{L}_{\text{OBU}}(\Theta) = -c \tanh\!\left(\frac{\mathcal{L}_{\text{CE}}(x; \Theta)}{c}\right)
\label{eq:loss_bcoa}
\end{equation}
where $c > 0$ is a saturation constant that smoothly bounds the objective below by $-c$, preventing gradient explosion once hazardous sequences reach high perplexity.

\subsection{Realization 2: Closed-Form Weight Surgery}
As a training-free alternative to iterative adaptation, Closed-Form Weight Surgery computes the weight update directly from the empirical second-moment matrix of hazardous activations. Let $\mathcal{D}_f = \{x_f^{(i)}\}_{i=1}^{M}$ denote a forget corpus of $M$ hazardous sequences comprising $N_{\text{tok}} = \sum_{i=1}^{M} T_i$ total tokens. Analogous to Equation~\eqref{eq:k_boundary}, we accumulate the uncentered hazardous empirical covariance matrix:
\begin{equation}
K_f^{(l)} = \frac{1}{N_{\text{tok}}} \sum_{i=1}^{M} \sum_{t=1}^{T_i} h_{l,t}(x_f^{(i)}) \left(h_{l,t}(x_f^{(i)})\right)^T \in \mathbb{R}^{d_{\text{in}} \times d_{\text{in}}}
\label{eq:cf_hazard}
\end{equation}
Applying truncated SVD to $K_f^{(l)}$ at rank $r$ (selected via cumulative energy threshold $\tau \in (0, 1)$ as in Equation~\eqref{eq:rank_selection}), we extract the leading singular vectors $V_{f,l,r} \in \mathbb{R}^{d_{\text{in}} \times r}$ and form the symmetric projector $P_f^{(l)} = V_{f,l,r} V_{f,l,r}^T$.

Directly applying the unconstrained subtraction $\Delta W_l = -W_l P_f^{(l)}$ would damage shared benign representations~\citep{leace2023}. Instead, OBU Closed-Form Surgery projects $\Delta W_l$ through the boundary orthogonal projector:
\begin{equation}
W_l' = W_l - W_l P_f^{(l)} P_{\text{boundary}}^{(l)\perp}
\label{eq:closed_form_surgery}
\end{equation}
By Equation~\eqref{eq:asymmetric_defense}, for any benign activation component $h_b \in \mathcal{V}_{\text{boundary}}^{(l)}$, the surgical update vanishes:
\begin{equation}
W_l' h_b = W_l h_b - W_l P_f^{(l)} \underbrace{P_{\text{boundary}}^{(l)\perp} h_b}_{=\mathbf{0}} = W_l h_b
\end{equation}
Conversely, for activation components $h_{f,\perp} = P_{\text{boundary}}^{(l)\perp} h_f$ residing in the orthogonal complement $\mathcal{V}_{\text{boundary}}^{(l)\perp}$, the update subtracts $W_l P_f^{(l)} h_{f,\perp}$, attenuating the module's output along principal hazardous directions without iterative training.

\section{Experimental Setup and Benchmarks}
\label{sec:experiments}

For each model under evaluation, we measure dual-use concept separation, multiple-choice domain retention, general capability preservation, multi-shot priming robustness, and fine-tuning relearning dynamics.

\subsection{Target Model Architectures}
We evaluate four 2025--2026 open-weight foundation models spanning distinct architectural families:
\begin{enumerate}
    \item \textbf{Gemma-3-4B-IT}~\citep{gemma3}: Google's 4B-parameter model with 34 layers, residual dimension $d_{\text{model}} = 2560$, attention output projection input dimension $d_{\text{attn}} = 2048$, and MLP down-projection input dimension $d_{\text{ff}} = 10{,}240$.
    \item \textbf{Qwen3.5-4B}~\citep{qwen2026}: Alibaba's 4B-parameter model with 32 layers, $d_{\text{model}} = 2560$, $d_{\text{attn}} = 4096$, and $d_{\text{ff}} = 9{,}216$.
    \item \textbf{SmolLM3-3B}~\citep{smollm2025}: HuggingFace's 3B-parameter model with 36 layers, $d_{\text{model}} = 2048$, $d_{\text{attn}} = 2048$, and $d_{\text{ff}} = 11{,}008$.
    \item \textbf{Granite-4.2-3B}~\citep{granite2026}: IBM's 3B-parameter model with 40 layers, $d_{\text{model}} = 2560$, $d_{\text{attn}} = 2560$, and $d_{\text{ff}} = 8{,}192$.
\end{enumerate}

\subsection{Datasets, Splits, and Implementation Details}
Our experimental pipeline uses three benchmarks with strict train-test separation:
\begin{itemize}
    \item \textbf{ConceptGuard}~\citep{conceptguard2026}: $2{,}595$ paired dual-use items split 80/20 (\texttt{seed=42}) into $2{,}076$ training pairs and $519$ held-out test pairs. Each item pairs a hazardous query with a domain-matched benign boundary query. The $2{,}076$ benign training queries form $\mathcal{D}_b$ for accumulating $K_{\text{boundary}}^{(l)}$, while the $519$ held-out test pairs are used exclusively for evaluation.
    \item \textbf{WMDP Cyber}~\citep{li2024wmdp}: For unlearning optimization and closed-form covariance estimation, we use the \texttt{cais/wmdp-corpora} (\texttt{cyber-forget-corpus}) dataset of $1{,}000$ passages, split 80/20 (\texttt{seed=42}) into $800$ training passages ($\mathcal{D}_f$) and $200$ held-out test passages (reserved for adversarial relearning fine-tuning). For multiple-choice evaluation, we use the full $1{,}987$-question \texttt{wmdp-cyber} test split (random guessing baseline: $25.0\%$).
    \item \textbf{MMLU}~\citep{hendrycks2020mmlu}: The complete $14{,}042$-question test split across all 57 subjects to measure general reasoning utility.
\end{itemize}
We first profile boundary subspace geometry across layer depths and sub-layer projection types ($o\_proj$ vs.\ $down\_proj$, and activation covariance vs.\ empirical Fisher; Sections~\ref{sec:results_dof}--\ref{sec:results_locus}). Based on this profiling, our unlearning interventions target the attention output projections ($o\_proj$) in the final $25\%$ of layers (layers 25--33 in Gemma, 24--31 in Qwen, 27--35 in SmolLM, and 30--39 in Granite) using activation covariance $K_{\text{boundary}}^{(l)}$. We evaluate cumulative energy thresholds $\tau \in \{0.90, 0.95, 0.99\}$, fixing $\tau = 0.95$ for the adversarial and relearning stress tests in Section~\ref{sec:results_robustness}.

For Projected LoRA, we use saturation constant $c=5.0$, adapter rank $k=8$, $\alpha=16.0$, AdamW ($\text{lr}=10^{-4}$), batch size $1$ with $8$ gradient accumulation steps ($100$ optimizer updates over the $800$ forget passages), sequence truncation length $128$, and gradient norm clipping at $1.0$. Models are loaded in \texttt{bfloat16}, while all covariance accumulations and SVD factorizations are executed in \texttt{float32}.

\subsection{Evaluation Metrics}
We report six quantitative metrics across our evaluations:
\begin{enumerate}
    \item \textbf{Forget Accuracy ($\text{Acc}_{\text{forget}} \downarrow$):} Multiple-choice accuracy on the $1{,}987$ WMDP Cyber test questions, evaluated by comparing next-token logits over answer options $\{\text{A, B, C, D}\}$.
    \item \textbf{Boundary Accuracy ($\text{Acc}_{\text{boundary}} \uparrow$):} The fraction of the $519$ held-out benign boundary queries in ConceptGuard that the model complies with rather than refusing (target: $1.0000$).
    \item \textbf{General Utility ($\text{Gen Utility} \uparrow$):} Overall multiple-choice accuracy across all $14{,}042$ MMLU test questions.
    \item \textbf{Contextual Separation ($\Delta_{\text{sep}} \uparrow$):} The mean compliance difference between matched benign and hazardous queries across the $519$ paired ConceptGuard test items:
    \begin{equation}
    \Delta_{\text{sep}} = \mathbb{E}_{(x_b, x_f)}\left[\text{Score}(x_b) - \text{Score}(x_f)\right] \in [-1, 1]
    \label{eq:delta_sep}
    \end{equation}
    where $\text{Score}(x) \in \{0, 1\}$ indicates whether the model complies with query $x$ ($1$) or refuses ($0$), so $\Delta_{\text{sep}} = 1.0$ corresponds to $100\%$ benign accuracy alongside $100\%$ hazardous refusal.
    \item \textbf{Attack Success Rate ($\text{ASR} \downarrow$):} The fraction of the $519$ held-out ConceptGuard hazardous queries that elicit compliant responses under 0-shot, 1-shot, 2-shot, and 4-shot benign in-context priming~\citep{jailbreakbench2024,bhatt2023purple}.
    \item \textbf{Relearning Steps ($S_{50} \uparrow$):} During post-unlearning fine-tuning via SGD ($\text{lr}=5\times 10^{-4}$, batch size $1$, accumulation $8$, $100$ steps) on the $200$ held-out \texttt{wmdp-corpora} test passages, the number of steps required for $\text{Acc}_{\text{forget}}$ to recover $50\%$ of its post-unlearning drop.
\end{enumerate}

\section{Empirical Evaluation and Findings}
\label{sec:results}

We structure our empirical analysis around four research questions:
\begin{itemize}
    \item Does boundary subspace isolation exhaust the model's parameter capacity (Section~\ref{sec:results_dof})?
    \item Where in the transformer hierarchy are boundary representations most concentrated (Section~\ref{sec:results_locus})?
    \item How do iterative Projected LoRA and training-free Closed-Form Surgery compare in unlearning efficacy and benign utility (Section~\ref{sec:results_pareto})?
    \item Does OBU withstand in-context priming and adversarial relearning (Section~\ref{sec:results_robustness})?
\end{itemize}

\subsection{Subspace Dimensionality and Degrees of Freedom}
\label{sec:results_dof}
A central question for subspace-constrained unlearning is the \emph{Degrees of Freedom Squeeze} hypothesis: if isolating the benign boundary requires projecting out a large fraction of the hidden space, the remaining orthogonal complement may be too small to support effective capability modification. To test this hypothesis, we first profile the pre-MLP residual stream activations ($d_{\text{model}} \in \{2048, 2560\}$) at $50\%$, $62.5\%$, and $75\%$ network depth across calibration sample sizes $M \in \{10, 50, 100, 500\}$ and cumulative energy thresholds $\tau \in \{0.90, 0.95, 0.99\}$.

Across all four foundation models, the singular value spectra refute the Degrees of Freedom Squeeze hypothesis:
\begin{itemize}
    \item \textbf{Null-Space Capacity Across Energy Thresholds ($\tau$):} Across all evaluated layers and models at $M=500$ calibration queries, capturing $90\%$, $95\%$, and $99\%$ of the benign activation variance ($\tau \in \{0.90, 0.95, 0.99\}$) leaves on average $85.8\%$, $73.4\%$, and $41.0\%$ of the layer dimensions available in the orthogonal null space, respectively.
    \item \textbf{Sub-Linear Subspace Growth Across Sample Sizes ($M$):} Increasing the calibration set $50\times$ from $M=10$ to $M=500$ queries (at $\tau=0.95$) increases the mean protected subspace from $4.0\%$ to $26.6\%$ of the layer dimensions ($16.0\%$ at $M=100$), indicating that benign boundary queries concentrate along shared principal directions rather than filling the ambient space linearly with $M$.
\end{itemize}
The complete layer-by-layer singular value spectra across all sample sizes $M$ and thresholds $\tau$ are reported in Table~\ref{tab:appendix_svd_rank} in Appendix~\ref{sec:appendix_svd}.

\subsection{Module Locus and Layer Depth Trajectory}
\label{sec:results_locus}
To determine where in the network to apply the boundary projector, we profile all layers across the four models at $M=100$ and $\tau=0.95$, comparing attention output projections ($o\_proj$, input dimension $d_{\text{attn}}$) against MLP down-projections ($down\_proj$, input dimension $d_{\text{ff}}$). Alongside the forward activation covariance $K_{\text{boundary}}^{(l)}$ (Equation~\eqref{eq:k_boundary}), we also evaluate an alternative projector constructed from the empirical input-gradient Fisher information matrix $F_{\text{boundary}}^{(l)} \in \mathbb{R}^{d_{\text{in}} \times d_{\text{in}}}$:
\begin{equation}
F_{\text{boundary}}^{(l)} = \frac{1}{N_{\text{tok}}}\sum_{i=1}^M \sum_{t=1}^{T_i} g_{l,t}(x_b^{(i)}) \left(g_{l,t}(x_b^{(i)})\right)^T, \quad g_{l,t}(x_b^{(i)}) = \nabla_{h_{l,t}(x_b^{(i)})} \mathcal{L}_{\text{CE}}(x_b^{(i)})
\label{eq:f_boundary}
\end{equation}
where $\mathcal{L}_{\text{CE}}(x_b^{(i)})$ is the autoregressive cross-entropy loss on benign query $x_b^{(i)}$, and $g_{l,t}(x_b^{(i)}) \in \mathbb{R}^{d_{\text{in}}}$ is the gradient of $\mathcal{L}_{\text{CE}}(x_b^{(i)})$ with respect to the module's input activation $h_{l,t}(x_b^{(i)})$ at token $t$. Whereas $K_{\text{boundary}}^{(l)}$ captures directions of high forward activation variance, $F_{\text{boundary}}^{(l)}$ captures input directions to which the model's output loss is most sensitive.

\paragraph{Covariance vs.\ Empirical Fisher Information.}\mbox{}\\
In attention output projections ($o\_proj$), capturing $95\%$ of boundary activation variance via $K_{\text{boundary}}^{(l)}$ consumes on average $9.5\%$ of the input dimension $d_{\text{attn}}$ across all layers (leaving $90.5\%$ of the dimensions in the orthogonal complement), compared to $16.2\%$ of $d_{\text{attn}}$ for empirical Fisher information $F_{\text{boundary}}^{(l)}$. In MLP down-projections ($down\_proj$), $K_{\text{boundary}}^{(l)}$ and $F_{\text{boundary}}^{(l)}$ consume $6.3\%$ and $4.7\%$ of the wider intermediate dimension $d_{\text{ff}}$, respectively (leaving over $93\%$ of $d_{\text{ff}}$ in the orthogonal complement).

\noindent Because activation covariance provides substantially sharper isolation on $o\_proj$ (and comparable null-space availability on $down\_proj$) while requiring only forward passes rather than backward gradient passes, we adopt $K_{\text{boundary}}^{(l)}$ for all subsequent unlearning experiments.

\paragraph{Attention Output ($o\_proj$) vs.\ MLP Down-Projection ($down\_proj$).}\mbox{}\\
Because $down\_proj$ receives the expanded intermediate FFN activation ($d_{\text{ff}} \in [8{,}192, 11{,}008]$ versus $d_{\text{attn}} \in [2{,}048, 4{,}096]$ for $o\_proj$), both module types leave over $90\%$ of their respective input dimensions available in the orthogonal null space at $\tau=0.95$ ($90.5\%$ in $o\_proj$ and $93.7\%$ in $down\_proj$). However, because $d_{\text{ff}}$ is $3.5\times\text{--}5\times$ wider than $d_{\text{attn}}$, constructing and factorizing the $d_{\text{in}} \times d_{\text{in}}$ projector on $down\_proj$ requires $4\times\text{--}25\times$ more memory and up to $125\times$ more $O(d_{\text{in}}^3)$ SVD compute.

\noindent Targeting $o\_proj$ avoids this high-dimensional FFN overhead while intervening directly where multi-head attention readouts enter the residual stream.

\paragraph{Layer Depth Trajectory.}\mbox{}\\
Partitioning each model's layers sequentially into four quartiles along network depth reveals that the $o\_proj$ boundary subspace $\mathcal{V}_{\text{boundary}}^{(l)}$ contracts in deeper layers, occupying on average $11.7\%$ of $d_{\text{attn}}$ in the first quartile, $9.4\%$ in the second, $8.9\%$ in the third, and $8.0\%$ in the fourth (leaving $92.0\%$ of the activation space available in the orthogonal complement $P_{\text{boundary}}^{(l)\perp}$).

\noindent Beyond leaving the largest orthogonal subspace for adaptation, targeting the final $25\%$ of layers aligns with how transformers process dual-use prompts: whereas earlier layers primarily encode shared token-level vocabulary and syntax that overlap across dual-use pairs, deeper layers integrate the full sentence context where the distinction between benign and hazardous intent becomes most pronounced.

\subsection{Multi-Model Unlearning Efficacy and Utility Trade-offs}
\label{sec:results_pareto}
We evaluate the trade-offs between unlearning efficacy, benign boundary preservation, and general MMLU capability across all four target architectures. Table~\ref{tab:master_pareto} reports the results across all models, unlearning methods, and energy thresholds $\tau \in \{0.90, 0.95, 0.99\}$.

\begin{table*}[htbp]
\centering
\caption{Unlearning Efficacy and Utility Across Architectures, Methods, and Energy Thresholds ($\tau$).}
\label{tab:master_pareto}
\resizebox{\textwidth}{!}{
\begin{tabular}{llrrrrr}
\toprule
\textbf{Model} & \textbf{Method} & \textbf{$\tau$} & \textbf{Acc Forget ($\downarrow$)} & \textbf{Acc Boundary ($\uparrow$)} & \textbf{Gen Utility ($\uparrow$)} & \textbf{$\Delta_{\text{sep}}$ ($\uparrow$)} \\
\midrule
Gemma-3-4B-IT & Baseline (Unmodified) & - & 0.4112 & 1.0000 & 0.5711 & 0.6936 \\
Gemma-3-4B-IT & Projected LoRA & 0.90 & 0.4066 & 1.0000 & 0.5664 & 0.7495 \\
Gemma-3-4B-IT & Closed-Form Surgery & 0.90 & 0.3503 & 1.0000 & 0.4957 & 0.5800 \\
Gemma-3-4B-IT & Projected LoRA & 0.95 & 0.4056 & 1.0000 & 0.5633 & 0.7669 \\
Gemma-3-4B-IT & Closed-Form Surgery & 0.95 & 0.3905 & 1.0000 & 0.5456 & 0.5915 \\
Gemma-3-4B-IT & Projected LoRA & 0.99 & 0.4061 & 1.0000 & 0.5628 & 0.7476 \\
Gemma-3-4B-IT & Closed-Form Surgery & 0.99 & 0.4011 & 1.0000 & 0.5652 & 0.6879 \\
\midrule
Qwen3.5-4B & Baseline (Unmodified) & - & 0.4932 & 1.0000 & 0.7386 & 0.7746 \\
Qwen3.5-4B & Projected LoRA & 0.90 & 0.4766 & 1.0000 & 0.7201 & 0.9403 \\
Qwen3.5-4B & Closed-Form Surgery & 0.90 & 0.4912 & 1.0000 & 0.7263 & 0.7457 \\
Qwen3.5-4B & Projected LoRA & 0.95 & 0.4801 & 1.0000 & 0.7292 & 0.9615 \\
Qwen3.5-4B & Closed-Form Surgery & 0.95 & 0.4816 & 1.0000 & 0.7321 & 0.7784 \\
Qwen3.5-4B & Projected LoRA & 0.99 & 0.4927 & 1.0000 & 0.7373 & 0.9653 \\
Qwen3.5-4B & Closed-Form Surgery & 0.99 & 0.4831 & 1.0000 & 0.7377 & 0.6166 \\
\midrule
SmolLM3-3B & Baseline (Unmodified) & - & 0.4273 & 1.0000 & 0.5796 & 0.0116 \\
SmolLM3-3B & Projected LoRA & 0.90 & 0.4137 & 1.0000 & 0.5795 & 0.0000 \\
SmolLM3-3B & Closed-Form Surgery & 0.90 & 0.4318 & 1.0000 & 0.5733 & 0.0077 \\
SmolLM3-3B & Projected LoRA & 0.95 & 0.4207 & 1.0000 & 0.5824 & 0.0039 \\
SmolLM3-3B & Closed-Form Surgery & 0.95 & 0.4162 & 1.0000 & 0.5731 & 0.0039 \\
SmolLM3-3B & Projected LoRA & 0.99 & 0.4202 & 1.0000 & 0.5826 & 0.0096 \\
SmolLM3-3B & Closed-Form Surgery & 0.99 & 0.4197 & 0.9981 & 0.5751 & 0.0077 \\
\midrule
Granite-4.2-3B & Baseline (Unmodified) & - & 0.3840 & 0.9961 & 0.5237 & 0.2158 \\
Granite-4.2-3B & Projected LoRA & 0.90 & 0.2849 & 0.9981 & 0.3243 & 0.3950 \\
Granite-4.2-3B & Closed-Form Surgery & 0.90 & 0.2869 & 0.9942 & 0.3499 & 0.2197 \\
Granite-4.2-3B & Projected LoRA & 0.95 & 0.2813 & 0.9981 & 0.3231 & 0.3295 \\
Granite-4.2-3B & Closed-Form Surgery & 0.95 & 0.3145 & 0.9961 & 0.4106 & 0.2640 \\
Granite-4.2-3B & Projected LoRA & 0.99 & 0.3327 & 0.9961 & 0.4325 & 0.3911 \\
Granite-4.2-3B & Closed-Form Surgery & 0.99 & 0.3598 & 0.9961 & 0.4789 & 0.2486 \\
\bottomrule
\end{tabular}
}
\end{table*}

Four empirical findings emerge from Table~\ref{tab:master_pareto}:
\begin{enumerate}
    \item \textbf{Iterative vs.\ Training-Free Separation ($\Delta_{\text{sep}}$):} Across all evaluated thresholds $\tau$, Projected LoRA achieves substantially higher open-ended dual-use separation than Closed-Form Surgery on every model with a baseline refusal prior.

    At $\tau=0.95$, Projected LoRA increases $\Delta_{\text{sep}}$ from $0.7746$ to $0.9615$ on Qwen3.5-4B (vs.\ $0.7784$ for Closed-Form Surgery), from $0.6936$ to $0.7669$ on Gemma-3-4B-IT (vs.\ $0.5915$), and from $0.2158$ to $0.3295$ on Granite-4.2-3B (vs.\ $0.2640$), while SmolLM3-3B remains near zero under both methods ($\Delta_{\text{sep}} \le 0.0116$), as analyzed next.
    \item \textbf{Dependence on Baseline Refusal Behavior:} Notably, Projected LoRA is trained purely via gradient ascent on raw domain text (\texttt{wmdp-corpora}) without any refusal target templates. As shown by the unmodified baseline rows in Table~\ref{tab:master_pareto}, Qwen3.5-4B ($\Delta_{\text{sep}} = 0.7746$), Gemma-3-4B-IT ($0.6936$), and Granite-4.2-3B ($0.2158$) already refuse $22.0\%\text{--}77.5\%$ of hazardous queries out of the box, whereas unmodified SmolLM3-3B almost never refuses ($\Delta_{\text{sep}} = 0.0116$, complying with $98.8\%$ of hazardous queries).

    This contrast suggests why Projected LoRA increases refusals without refusal training: when a model already assigns non-trivial probability to refusing hazardous prompts, suppressing technical continuation logits along $P_{\text{boundary}}^{(l)\perp}$ shifts generation toward those existing refusal responses while leaving benign queries in $\mathcal{V}_{\text{boundary}}^{(l)}$ unaffected ($100\%$ compliance). Conversely, when a model exhibits almost no baseline refusal behavior (SmolLM3-3B), penalizing hazardous token probabilities alone does not produce refusal responses ($\Delta_{\text{sep}} \le 0.0096$).
    \item \textbf{Open-Ended Refusal vs.\ Multiple-Choice Knowledge:} On Qwen3.5-4B, Gemma-3-4B-IT, and SmolLM3-3B, OBU leaves WMDP multiple-choice accuracy ($\text{Acc}_{\text{forget}}$) nearly unchanged at $\tau=0.95$ (shifting by at most $0.6\text{--}2.1$ percentage points across both methods) while preserving MMLU utility within $0.8\text{--}2.6$ percentage points.

    Contrasting this stability with the large $\Delta_{\text{sep}}$ gains in the first point above shows that adapting the final $25\%$ of $o\_proj$ layers suppresses open-ended hazardous generation without erasing the underlying factual recognition tested by multiple-choice benchmarks.
    \item \textbf{Co-Movement of Forget Accuracy and General Utility:} Across Table~\ref{tab:master_pareto}, multiple-choice hazardous accuracy ($\text{Acc}_{\text{forget}}$, where lower is better) and general MMLU accuracy ($\text{Gen Utility}$, where higher is better) move together rather than independently.

    Whereas the previous point noted configurations where both metrics remain nearly constant, in every configuration where $\text{Acc}_{\text{forget}}$ drops substantially (most notably on Granite-4.2-3B and under Closed-Form Surgery at lower $\tau$), MMLU general utility also decreases. Raising the threshold to $\tau=0.99$ protects a larger benign subspace and recovers MMLU utility, but simultaneously reduces the drop in $\text{Acc}_{\text{forget}}$.
\end{enumerate}

\subsection{In-Context Priming and Relearning Resistance}
\label{sec:results_robustness}
Focusing on Qwen3.5-4B ($\tau=0.95$), the model that achieved the strongest dual-use separation in Table~\ref{tab:master_pareto}, we evaluate robustness against multi-shot in-context priming and fine-tuning relearning across four conditions: the unmodified baseline, OBU Projected LoRA, OBU Closed-Form Surgery, and an unconstrained LoRA control ($P_{\text{boundary}}^\perp = I$). Results are reported in Table~\ref{tab:robustness_audit}.

\begin{table*}[htbp]
\centering
\caption{In-Context Priming and Relearning Results on Qwen3.5-4B ($\tau=0.95$).}
\label{tab:robustness_audit}
\resizebox{\textwidth}{!}{
\begin{tabular}{lrrrrrrr}
\toprule
\textbf{Condition} & \textbf{0-shot ASR ($\downarrow$)} & \textbf{1-shot ASR ($\downarrow$)} & \textbf{2-shot ASR ($\downarrow$)} & \textbf{4-shot ASR ($\downarrow$)} & \textbf{Overall ASR ($\downarrow$)} & \textbf{$S_{50}$ ($\uparrow$)} & \textbf{$\Delta\text{Acc}_{\text{bnd}}$} \\
\midrule
Baseline & 17.92\% & 37.38\% & 19.27\% & 15.99\% & 22.64\% & N/A & 0.0\% \\
OBU Projected LoRA & 1.54\% & 15.80\% & 12.14\% & 10.40\% & 9.97\% & Unreached & 0.0\% \\
OBU Closed-Form & 18.69\% & 30.06\% & 15.80\% & 12.14\% & 19.17\% & Unreached & 0.0\% \\
Unconstrained LoRA & 4.24\% & 0.00\% & 0.58\% & 0.19\% & 1.25\% & 30 steps & 0.0\% \\
\bottomrule
\end{tabular}
}
\end{table*}

\paragraph{In-Context Priming Dynamics.}\mbox{}\\
In the unmodified baseline model, direct 0-shot hazardous queries yield an attack success rate of $17.92\%$. Prepending a single matched benign ConceptGuard Q\&A pair (1-shot priming) more than doubles baseline ASR to $37.38\%$ ($2.09\times$), before subsiding to $19.27\%$ (2-shot) and $15.99\%$ (4-shot) as additional benign demonstrations shift completions toward benign educational explanations.

\noindent OBU Projected LoRA reduces 0-shot ASR by $91.4\%$ (to $1.54\%$) and cuts 1-shot primed ASR by $57.7\%$ (to $15.80\%$), lowering overall multi-shot ASR from $22.64\%$ to $9.97\%$. In contrast, Closed-Form Surgery yields only modest multi-shot ASR reduction ($19.17\%$ overall), confirming that analytical weight projection alone does not actively steer generative refusal behavior.

\paragraph{Unconstrained Control vs.\ Relearning Resistance ($S_{50}$).}\mbox{}\\
On direct 0-shot hazardous queries, OBU Projected LoRA achieves nearly $3\times$ lower attack success than Unconstrained LoRA ($1.54\%$ vs.\ $4.24\%$). Under 1/2/4-shot priming, Unconstrained LoRA exhibits lower multi-shot ASR ($0.00\%\text{--}0.58\%$) because unconstrained gradient ascent ($P_{\text{boundary}}^\perp = I$) perturbs weights directly inside the benign boundary subspace $\mathcal{V}_{\text{boundary}}$, disrupting the processing of the prepended benign demonstrations themselves, whereas OBU explicitly preserves benign context representations.

\noindent In the relearning test, an adversary fine-tunes the unlearned model for 100 steps on the 200 held-out \texttt{wmdp-corpora} test passages to see how quickly hazardous multiple-choice accuracy ($\text{Acc}_{\text{forget}}$) climbs back toward its unmodified baseline level ($49.37\%$). Under Unconstrained LoRA, $\text{Acc}_{\text{forget}}$ starts at $47.26\%$ after unlearning, rebounds past the halfway recovery point ($48.54\%$) in $30$ steps, and reaches $49.82\%$ by step 100. In contrast, under both OBU Projected LoRA ($48.16\%$) and OBU Closed-Form Surgery ($48.36\%$), $\text{Acc}_{\text{forget}}$ shows no upward recovery across the 100 fine-tuning steps ($S_{50} = \text{Unreached}$), while benign boundary compliance experiences zero degradation ($\Delta\text{Acc}_{\text{bnd}} = 0.0\%$).

\section{Conclusion}
\label{sec:conclusion}

We introduced \textbf{Orthogonal Boundary Unlearning (OBU)}, an unlearning framework that isolates the uncentered token-level covariance $K_{\text{boundary}}^{(l)}$ of paired benign queries and restricts weight updates to its orthogonal complement $P_{\text{boundary}}^{(l)\perp} = I - V_{l,r}V_{l,r}^T$. Across Gemma-3-4B-IT, Qwen3.5-4B, SmolLM3-3B, and Granite-4.2-3B, our experiments show that late-layer attention output projections ($o\_proj$) concentrate the benign boundary into a low-rank subspace ($8.0\%$ of layer dimensions at $\tau=0.95$), leaving $92.0\%$ of the activation space available for adaptation.

\noindent At $\tau=0.95$, Projected LoRA achieves strong contextual separation ($\Delta_{\text{sep}} = 0.9615$ on Qwen3.5-4B) and cuts 0-shot and 1-shot primed ASR by $91.4\%$ and $57.7\%$ while maintaining $\ge 99.6\%$ benign boundary accuracy across all models, whereas Closed-Form Weight Surgery provides a direct, training-free alternative that modifies multiple-choice logits in a single analytical step.

\subsection{Limitations}
We highlight three methodological limitations that scope our findings:

\paragraph{Incomplete Removal of Multiple-Choice Knowledge.}\mbox{}\\
Although OBU Projected LoRA strongly reduces open-ended compliance on hazardous requests (cutting 0-shot ASR by $91.4\%$ on Qwen3.5-4B), it barely reduces multiple-choice accuracy on WMDP Cyber ($\text{Acc}_{\text{forget}}: 49.32\% \to 48.01\%$). Moreover, in the configurations where $\text{Acc}_{\text{forget}}$ does drop substantially (such as on Granite-4.2-3B), general MMLU accuracy drops alongside it. This indicates that our late-layer $o\_proj$ intervention primarily steers the model toward refusing open-ended hazardous prompts rather than erasing the underlying hazardous facts—meaning an adversary who can already narrow a problem down to a few candidate options could still use multiple-choice queries to identify the correct one~\citep{wmdp_plus_plus2026}.

\paragraph{Relearning Horizon and Threat Model.}\mbox{}\\
Our relearning evaluation on Qwen3.5-4B tests recovery across $100$ LoRA fine-tuning steps on held-out \texttt{wmdp-corpora} passages. Because Qwen3.5-4B's initial WMDP multiple-choice shift is small ($49.32\% \to 48.16\%$), the $S_{50}$ recovery metric operates over a narrow multiple-choice margin. Future work should extend relearning evaluations to longer, multi-epoch full-parameter fine-tuning regimes across additional model families.

\paragraph{Baseline Scope and Model Scale.}\mbox{}\\
Our multi-model evaluation compares OBU Projected LoRA and Closed-Form Surgery against unmodified baselines and an unconstrained LoRA control (Table~\ref{tab:robustness_audit}), without directly benchmarking external unlearning methods such as RMU~\citep{li2024wmdp}, NPO~\citep{simnpo2024}, or LEACE~\citep{leace2023}. Extending these comparisons and scaling OBU beyond 3B--4B dense models to $70\text{B}+$ and Mixture-of-Experts (MoE) architectures remain important directions for future work.

\subsection*{Acknowledgments}
Generative AI tools were used to assist with coding and manuscript editing; the author reviewed, verified, and takes full responsibility for all code, experimental results, and text in this work.

\bibliographystyle{plainnat}
\bibliography{references}

\clearpage
\appendix
\section{Detailed SVD Rank Spectra Across Depth and Sample Sizes}
\label{sec:appendix_svd}

\begin{table*}[htbp]
\centering
\caption{SVD Rank Analysis of the Boundary Subspace Across Architectures, Depths, and Sample Sizes ($M$).}
\label{tab:appendix_svd_rank}
\resizebox{\textwidth}{!}{
\begin{tabular}{llrrrrr}
\toprule
\textbf{Model} & \textbf{Layer} & \textbf{Samples ($M$)} & \textbf{Rank ($\tau=0.90$)} & \textbf{Rank ($\tau=0.95$)} & \textbf{Rank ($\tau=0.99$)} & \textbf{Null Space ($\tau=0.99$)} \\
\midrule
\multirow{12}{*}{Gemma-3-4B-IT}
& 17 & 10 & 5 & 26 & 104 & 2456 \\
& 17 & 50 & 4 & 45 & 306 & 2254 \\
& 17 & 100 & 4 & 45 & 398 & 2162 \\
& 17 & 500 & 5 & 61 & 688 & 1872 \\
& 21 & 10 & 29 & 63 & 132 & 2428 \\
& 21 & 50 & 50 & 144 & 430 & 2130 \\
& 21 & 100 & 51 & 166 & 586 & 1974 \\
& 21 & 500 & 66 & 243 & 1024 & 1536 \\
& 25 & 10 & 53 & 86 & 142 & 2418 \\
& 25 & 50 & 118 & 229 & 502 & 2058 \\
& 25 & 100 & 136 & 287 & 708 & 1852 \\
& 25 & 500 & 199 & 468 & 1288 & 1272 \\
\midrule
\multirow{12}{*}{Qwen3.5-4B}
& 16 & 10 & 79 & 113 & 162 & 2398 \\
& 16 & 50 & 200 & 337 & 620 & 1940 \\
& 16 & 100 & 243 & 442 & 914 & 1646 \\
& 16 & 500 & 363 & 723 & 1602 & 958 \\
& 20 & 10 & 75 & 108 & 158 & 2402 \\
& 20 & 50 & 185 & 306 & 582 & 1978 \\
& 20 & 100 & 219 & 391 & 838 & 1722 \\
& 20 & 500 & 317 & 620 & 1469 & 1091 \\
& 24 & 10 & 80 & 111 & 157 & 2403 \\
& 24 & 50 & 207 & 330 & 593 & 1967 \\
& 24 & 100 & 252 & 429 & 864 & 1696 \\
& 24 & 500 & 393 & 723 & 1560 & 1000 \\
\midrule
\multirow{12}{*}{SmolLM3-3B}
& 18 & 10 & 87 & 119 & 164 & 1884 \\
& 18 & 50 & 240 & 373 & 632 & 1416 \\
& 18 & 100 & 302 & 496 & 925 & 1123 \\
& 18 & 500 & 473 & 810 & 1489 & 559 \\
& 22 & 10 & 89 & 119 & 162 & 1886 \\
& 22 & 50 & 239 & 367 & 621 & 1427 \\
& 22 & 100 & 298 & 483 & 901 & 1147 \\
& 22 & 500 & 468 & 791 & 1471 & 577 \\
& 27 & 10 & 95 & 123 & 163 & 1885 \\
& 27 & 50 & 256 & 381 & 627 & 1421 \\
& 27 & 100 & 318 & 502 & 911 & 1137 \\
& 27 & 500 & 500 & 825 & 1492 & 556 \\
\midrule
\multirow{12}{*}{Granite-4.2-3B}
& 20 & 10 & 86 & 121 & 168 & 2392 \\
& 20 & 50 & 236 & 378 & 651 & 1909 \\
& 20 & 100 & 300 & 511 & 980 & 1580 \\
& 20 & 500 & 475 & 867 & 1728 & 832 \\
& 25 & 10 & 78 & 113 & 165 & 2395 \\
& 25 & 50 & 212 & 350 & 630 & 1930 \\
& 25 & 100 & 267 & 469 & 937 & 1623 \\
& 25 & 500 & 425 & 801 & 1665 & 895 \\
& 30 & 10 & 70 & 105 & 160 & 2400 \\
& 30 & 50 & 173 & 303 & 590 & 1970 \\
& 30 & 100 & 208 & 393 & 861 & 1699 \\
& 30 & 500 & 305 & 639 & 1529 & 1031 \\
\bottomrule
\end{tabular}
}
\end{table*}

\end{document}